# Lexical Access for Speech Understanding using Minimum Message Length Encoding

**Ian Thomas, Ingrid Zukerman, Jonathan Oliver, David Albrecht**
Department of Computer Science
Monash University
Clayton, Victoria 3168, AUSTRALIA
{iant,ingrid,jono,dwa}@cs.monash.edu.au

**Bhavani Raskutti**
Artificial Intelligence Section
Telstra Research Laboratories
Clayton, Victoria 3168, AUSTRALIA
b.raskutti@trl.oz.au

## Abstract

The *Lexical Access Problem* consists of determining the intended sequence of words corresponding to an input sequence of phonemes (basic speech sounds) that come from a low-level phoneme recognizer. In this paper we present an information-theoretic approach based on the Minimum Message Length Criterion for solving the Lexical Access Problem. We model sentences using phoneme realizations seen in training, and word and part-of-speech information obtained from text corpora. We show results on multiple-speaker, continuous, read speech and discuss a heuristic using equivalence classes of similar sounding words which speeds up the recognition process without significant deterioration in recognition accuracy.

## 1 INTRODUCTION

The *Lexical Access Problem* consists of determining the sequence of words that corresponds to an input sequence of phonemes (basic speech sounds). A lexical access component is a major part of a speech recognition system which discovers the sentences (composed of words from a lexicon) that correspond to speech signals.

If the sequences of phonemes we are given correspond precisely to words in the lexicon, we can imagine lexical access as a table lookup process, i.e., we simply select the words in the lexicon that have the same canonical phoneme sequences as the sequences in the input. However, in reality, the process of matching input phoneme sequences to words in a lexicon is more difficult, as these sequences may have extra or missing phonemes, and some phonemes may have been transcribed incorrectly. These insertions, deletions and substitutions may be due to (1) mis-recognition, through poor equipment, bad recording conditions, or poorly trained phoneme models; or (2) mis-pronunciation, where a speaker has said a word in a different way to the lexicon's canonical versions of that word. Mis-pronunciation is caused primarily by different dialects and accents. For example, the word "another" may be pronounced as several different sequences of phonemes:

| | | | | | | |
|---|---|---|---|---|---|---|
| Lexicon entry[1]: | *another* | ax | n | ah | dh | axr |
| Data: | | er | n | ah | dh | axr |
| | | | en | ah | dh | axr |
| | | ax | n | ah | dh | er |

Moreover, lexical access on entire sentences has the added uncertainty of knowing neither the number of words in the sentence nor the start and end points for each word in the sentence. Determining the boundaries of words in continuous speech is an extremely difficult task due to the above mentioned insertions, deletions and substitutions, and because often there are no word boundaries in the speech signal as a result of co-articulation – the "slurring" of sounds in continuous speech. The co-articulation effect can occur in any phonemes in an utterance, causing phonemes to be affected by their surrounding phonemes, which in turn leads to further mis-recognitions. In practical systems, lexical access is performed by attempting different partitions of the input phoneme string, and then looking for the optimal fitting of postulated words over the different phoneme subsequences resulting from these partitions (Myers and Rabiner, 1981, Lee and Rabiner, 1989).

In this paper, we describe a lexical access method for speech understanding based on the *Minimum Message Length (MML)* principle (Wallace and Freeman, 1987). This principle provides us with a uniform and incremental framework for applying information from different sources, such as a language model, lexicon and prosodic information, to the lexical access problem.

In Section 2 we discuss related research. We then describe a method for evaluating a sentence in an information-theoretic framework, and describe the search through the set of possible sentences corresponding to a given input. We conclude by discussing results obtained on a test set of sentences.

## 2 RELATED RESEARCH

In our study, we have as input a string of phoneme symbols and hypothesize the sentence that corresponds to it. These symbols would be the best phoneme candidates from the

[1] Phonemes are described using ARPAbet symbols, which are used in the TIMIT corpus (Fisher *et al.*, 1986). ARPAbet is a typewritten version of the standard International Phonetic Alphabet.

```
Withdraw all phony accusations at once.
withdraw                       |all      |phony             |
W    ih   dh   D    R    ao   |ao   L   |F    OW   n    Y   |
W    ix   th   D    R    aa   |aa   L   |F    OW   nx   Y   |

accusations                                          |at            |once                    |
AE   K    Y    uw   Z    EY   SH   ix   n    Z       |-    ae   T   |W    AH   N    -    S    |
AE   K    Y    ux   Z    EY   SH   en   epi  Z       |q    eh   T   |W    AH   N    t    S    |
```

Figure 1: A typical TIMIT sentence aligned with canonical phonemes from words in the lexicon.

output of a phoneme recognizer that works directly on the speech signal (Grayden and Scordilis, 1994). In most high-performance speech recognition systems, the lexical access operation is integrated with phoneme recognition and language modeling in the form of a graph search through, in effect, a massive Hidden Markov Model (HMM) for sentences, e.g., (Gauvain *et al.*, 1995, Jeanrenaud *et al.*, 1995). In our research, we isolate two parts of this process, namely word hypothesis generation and language modeling, in order to study their effect on the lexical access problem.

Efforts in lexical access have often followed the hypothesize-and-test paradigm, where the waveform corresponding to a word is partitioned and each segment labelled according to relatively reliable information extracted from the signal (such as whether the segment is voiced or unvoiced). Word candidates that fit a given string of labels undergo a more detailed and time-consuming analysis to determine the candidate that best matches the waveform (Fissore *et al.*, 1988). We use a similar method in our system. On a sentence level, the analysis of the sentence waveform to postulate individual words is often guided by detected word boundaries (Murveit *et al.*, 1987). Our method would be able to make more informed and quicker decisions if it had such information regarding word boundaries, but it can proceed in the absence of such information.

HMMs for sentences are normally built from smaller embedded HMMs for words (Rabiner and Juang, 1993). We use a mixture of an order two and order three Markov chain for the language model, but our word model is based on phonetic similarity, and can be generated directly from the training data without algorithms such as the Forward-backward Algorithm (Baum, 1972). In addition, unlike other methods for encoding acoustic and language models, e.g., (Béchet *et al.*, 1994, Zue and Lamel, 1986), we take an information-theoretic approach to representing data. Our work is similar to recent work on handwriting recognition (Bouchaffra *et al.*, 1997). But we use $N$-grams to estimate the parameters of the language model, while they use informative Dirichlet priors for estimating the same parameters.

## 3 DOMAIN OF STUDY

Our study was done on the TIMIT corpus (Fisher *et al.*, 1986), which is a collection of American-English read sentences with correct time-aligned acoustic-phonetic and orthographic (word-aligned) transcriptions. The data set contains 3696 sentences spoken by 462 speakers from 8 different dialect divisions across the United States. Each speaker says five phonetically-compact sentences and three phonetically-diverse sentences to give a good coverage of the phonemes in the language. The sentences were recorded using a high-quality, headset-mounted microphone in a noise-isolated room, and speakers were instructed to read prompts in a "natural" voice. This training set generated 18854 training words, with a total lexicon of 6236 distinct words.

An example of a TIMIT sentence aligned with canonical phonemes from words in the lexicon is given in Figure 1. The first row shows the words of the utterance, the second row the phonemes from the lexicon, and the third row the actual phonemes spoken by one of the TIMIT speakers.

The probability of a sentence $W*$ can be calculated as

$$P(W*) = P(w_1 w_2 w_3 \ldots w_n) = P(w_1)P(w_2|w_1)P(w_3|w_1 w_2)\ldots P(w_n|w_1 w_2 \ldots w_{n-1}).$$

However, it is not feasible to reliably calculate the conditional probabilities $P(w_j|w_1 \ldots w_{j-1})$ for all words and all partial sentence lengths, therefore we estimate such probabilities by using an $N$-gram:

$$P(w_j|w_1 w_2 \ldots w_{j-1}) \approx P(w_j|w_{j-N+1} \ldots w_{j-1}).$$

$N$-grams over part-of-speech symbols and words can be collected from training texts.

One difficulty encountered is that although TIMIT is excellent for providing many different examples of words in many different contexts, it is poor training data for a language model. This is due to the relatively small number of different sentences in the training corpus; but more particularly because the sentences are designed to be diverse and unusual. A language model trained on unusual sentences is unlikely to be generally useful, and its usefulness is even more questionable in the recognition of sentences from an unusual test set, as is the case in TIMIT. These facts make TIMIT a poor model for common English.

This problem was handled using TIMIT for the mappings between words and phoneme strings, and using a set of classics texts[1] to extract word and part-of-speech $N$-grams required for the language model (Section 4.2.1). This data provided many more co-locations of words and part-of-speech symbols and more varied ones than was possible with the TIMIT data: approximately 17000 distinct part-of-speech trigrams, compared with only 3000 from the TIMIT data. A small training set would cause more zero-frequency $N$-grams to be encountered during testing than a large training set, forcing us to use lower-context $N$-grams to estimate the probabilities of the components of higher-context $N$- grams. This zero-frequency effect should be reduced by using a larger training set, but such a training set forces us to store far more $N$-grams that may ever be encountered in the input to the system. Finally, using different training sets produces different probabilities for the $N$-grams; we would hope that the chosen training set approximates the "correct" probabilities for any sentence in general English.

[1]These publicly available texts include a variety of works such as Virgil's *Aeneid* and Emily Bronte's *Wuthering Heights*.

# 4 METHOD

Our approach to the lexical selection problem is based on the *Minimum Message Length (MML)* criterion (Wallace and Freeman, 1987). According to this criterion we imagine sending to a receiver the shortest possible message that describes a sequence of input phonemes. Now, this message may be composed of the given phoneme sequence or of a sequence of words that correspond to this phoneme sequence, i.e., a sentence. We postulate that a message that encodes a sequence of phonemes as a sentence will be shorter than a message that encodes it directly. Further, we postulate that the message describing the intended sentence will be among the sentences of shortest message length (hopefully the shortest). Thus, in finding the sequence of words that yields the shortest message given an input phoneme sequence we will have solved the lexical access problem. To find this sequence of words, we perform a search through a set of likely sentences, evaluating each sentence to find its message length.

## 4.1 MINIMUM MESSAGE LENGTH ENCODING

We use a message of two parts to describe a sequence of phonemes: (1) *model description segment* that describes the word sequence that the string of phonemes represents; and (2) *object description segment* that describes for each phoneme in the input string the deviation from its corresponding phoneme in the set of phonemes predicted by the word sequence. For example, the difference between "dh ae r" (a possible realization of "there") and the input sequence "dh ax r" is the substitution of "ax" for "ae".

The message can be thought of as an explanation of the data. The first segment is a theory about the phonemes based on the words postulated for the sentence, and the second a description of the actual phonemes in terms of the model phonemes in the postulated words. If the theory explains the data well, then the data description will be short. If the theory is poor, then the data description will be longer, causing a longer message length. The "best" theory is the one with the shortest total message length. A complicated theory segment will not necessarily cause a long total message length, nor will a short theory automatically cause a short message length. The final length of a message depends both on the length of the theory and on how well the theory describes the data.

The Minimum Message Length criterion is derived from Bayes Theorem:

$$P(H\&D) = P(H) \times P(D|H),$$

where $H$ is the hypothesis and $D$ is the data.

An optimal code for an event $E$ with probability $P(E)$ has message length $ML(E) = -\log_2(P(E))$. Hence, the message length for a hypothesis given the data is:

$$ML(H\&D) = ML(H) + ML(D|H),$$

which corresponds to the two parts of the message. The minimization of $ML(H\&D)$ is the criterion for model selection. The relationship between MML and Bayesian posterior maximization is discussed in (Oliver and Baxter, 1994).

## 4.2 EVALUATION OF A SENTENCE

The description of a sequence of phonemes is made up of two main parts: (1) Language Model, which describes the words; and (2) Phoneme Realization Difference, which describes the phonemes corresponding to this model, with an error function that describes the difference between the phonemes in the language model and the actual phonemes.

### 4.2.1 Language Model

Here we have the task of encoding the actual words that make up our sentence. Since a "sensible" sounding sentence would be better than a sentence composed of unrelated words, we require an encoding that will describe sensible sentences in as few bits as possible. To express how sensible a combination of words in a sentence is, we take into account the syntactic role of the words in this combination as well as the actual usage of these words. The former is done by preferring frequent part-of-speech combinations, e.g., an article followed by a noun, to infrequent ones, e.g., an article followed by another article; and the latter by preferring common word combinations.

Let $W* = w_1, w_2, \ldots, w_n$ be a sentence where word $w_i$ has been instantiated with part-of-speech $PoS_i$ (a word could correspond to different part-of-speech symbols).

$$\begin{array}{ccccc} W* & w_1 & w_2 & \ldots & w_n \\ & PoS_1 & PoS_2 & \ldots & PoS_n. \end{array}$$

We take the probability of sentence $W*$ as:

$$P(W*) = \prod_{i=1}^{n} P(w_i, PoS_i|\text{words before position } i).$$

At present, we use trigrams for parts-of-speech and bigrams for words, yielding:

$$\begin{aligned} P(W*) &= \prod_{i=1}^{n} P(w_i, PoS_i|w_{i-1}, PoS_{i-1}, PoS_{i-2}) \\ &= \prod_{i=1}^{n} P(w_i|PoS_i, w_{i-1}, PoS_{i-1}, PoS_{i-2}) \\ &\quad \times P(PoS_i|w_{i-1}, PoS_{i-1}, PoS_{i-2}). \end{aligned}$$

To reduce the complexity of the problem we assume that given the part-of-speech of word $w_i$, and given word $w_{i-1}$, word $w_i$ is conditionally independent of the previous parts-of-speech $PoS_{i-1}$ and $PoS_{i-2}$:

$$P(w_i|PoS_i, w_{i-1}, PoS_{i-1}, PoS_{i-2}) = P(w_i|PoS_i, w_{i-1}),$$

and that given the previous parts-of-speech $PoS_{i-1}$ and $PoS_{i-2}$, the part-of-speech $PoS_i$ is conditionally independent of the previous word $w_{i-1}$:

$$P(PoS_i|w_{i-1}, PoS_{i-1}, PoS_{i-2}) = P(PoS_i|PoS_{i-1}, PoS_{i-2}).$$

Using these conditional independence assumptions, we get:

$$P(W*) = \prod_{i=1}^{n} P(w_i|PoS_i, w_{i-1}) \times P(PoS_i|PoS_{i-1}, PoS_{i-2}).$$

The part-of-speech trigrams are estimated from frequencies of a training corpus according to the formula

$$\hat{P}(PoS_i|PoS_{i-1}, PoS_{i-2}) = \frac{F(PoS_i, PoS_{i-1}, PoS_{i-2})}{F(PoS_{i-1}, PoS_{i-2})},$$

where $F(x)$ is the frequency of $x$ seen in the training data. A similar formula is used to estimate $P(w_i|PoS_i, w_{i-1})$.

The larger the number of different symbols used in an $N$-gram and the larger the value for $N$, the less likely it is that a given set of training data will have instances of every $N$-gram. We handle the *Zero Frequency Problem* by using a back-off procedure to a lower context $N$-gram. This back-off is indicated by an escape code of low probability that is added to the code space of the $N$-gram (Witten and Bell, 1991). For example, if we have no instances of a particular part-of-speech trigram, we indicate this using the escape code, and then use a bigram. For words, we back-off from bigrams to unigrams (there is assumed to be one instance of every word in the lexicon in the unigram set).

To illustrate the above process, consider the following sentence fragment:

| | | | | |
|---|---|---|---|---|
| **PoS symbols:** | article | adjective | adjective | noun |
| **Words:** | *the* | *quick* | *brown* | *fox* |

which is encoded as the product of the following probabilities:

| | |
|---|---|
| $P(the\|eos, the = \text{art})$ | $P(\text{art}\|eos1, eos2)$[1] |
| $P(quick\|the, quick = \text{adj})$ | $P(\text{adj}\|\text{art}, eos1)$ |
| $P(brown\|quick, brown = \text{adj})$ | $P(\text{adj}\|\text{adj}, \text{art})$ |
| $P(fox\|brown, fox = \text{noun})$ | $P(\text{noun}\|\text{adj}, \text{adj})$ |

A search through all possible sentences with this evaluation scheme would tend to generate the most common possible sentences, without regard to how closely they match our actual phoneme input. The second part of our model tempers this effect by incorporating the actual phoneme data.

### 4.2.2 Phoneme Realization Difference

We have a set of words and a partition of the stream of phonemes into the words in the set. We have described the words; now we can make a hypothesis about the phonemes that correspond to these words, and we can measure how close reality is to our hypothesis.

Initially, we use the training set to generate for each word in the lexicon a set of possible phoneme realizations together with the frequency of each realization. We use an edit distance algorithm (Sandoff and Kruskal, 1983) to work out the realization of a lexicon word which is closest to our segment of actual phonemes. This algorithm yields an optimal alignment of the input phonemes with those in lexicon words using weights for insertions, substitutions and deletions which were obtained from training data (Thomas *et al.*, forthcoming). For example, phonetically similar substitutions are given a low cost (e.g., vowels for vowels).

As stated above, in the framework of the MML principle, the central idea is that the message sent to a receiver must contain enough information so that the receiver can reconstruct the input phonemes. We postulate that the shortest message will be composed of the lexicon word whose realization best matches the given input, the realization in question, and a record of the operations required to transform this realization into the actual input phonemes. Recall that the language model is being used to send the lexicon word. Hence, at this stage we only need to send the other two components. For example, suppose we wish to send the phoneme string " ax n dx q er", and we hypothesize that it corresponds to the word "another". The following realizations of this word have been seen in training:

| Frequency | Word | Phonemes | | | | | |
|---|---|---|---|---|---|---|---|
| 3 | $another_1$ | ix | n | ah | dh | axr | |
| 3 | $another_2$ | ax | n | ah | dh | er | |
| 2 | $another_3$ | q | ax | n | ah | dh | axr |
| 1 | $another_4$ | q | ax | n | ah | dh | er |
| 1 | $another_5$ | q | ax | n | ah | dh | ax |
| 1 | $another_6$ | ix | nx | ah | dh | uh | |
| 1 | $another_7$ | er | n | ah | dh | axr | |

In this example, the edit distance algorithm chooses the second realization, and it takes

$$-\log_2 \frac{\text{Frequency of } another_2}{\sum_i \text{Frequency of } another_i} = -\log_2(3/12)$$

bits to indicate to the receiver which phoneme realization we are going to use. However, the input phonemes do not correspond exactly to this realization, so we calculate an optimal alignment between our hypothesis phonemes and the input phonemes. For instance, for the second realization, the optimal alignment is

| | | | | | | |
|---|---|---|---|---|---|---|
| "$another_2$" | ax | n | ah | dh | – | er |
| input | ax | n | – | dx | q | er |

where "dh" and "dx" are aligned because they are phonetically similar (they have a low substitution cost). Since we have already specified which of the phoneme realizations to use, we have already encoded the hypothesis phonemes (the top line). We must now send the actual phonemes (bottom line), which are sent as a sequence of insertions, deletions and substitutions. This is performed in two stages: (1) sending insertions, and (2) sending the rest of the operations.

**Sending insertions**. Insertions are special because they increase the number of operations in the alignment, and therefore have to be specified using position indicators.

To send insertions, we first specify how many insertions there are in the alignment, so that the length of the part of the message that describes which phonemes were inserted can be determined. This information takes $-\log_2(P(N \text{ insertions}))$ bits to transmit (in our example only one insertion was performed, namely "q" between "dx" and "er").

[1] All sentences start with *eos*1 (*end of string*) and *eos*2 part-of-speech tags, and an *eos* word.

The probability of having $N$ insertions is estimated from the training corpus as follows: 10% of the database of realizations seen in training is removed; each realization in this subset is optimally aligned with the remaining realizations for the same word in the database; and the number of insertions in the closest alignment is recorded. This process is repeated by removing a different 10% of the database until all phoneme realizations in the original database have been used once.

Next, we need to specify what the inserted phonemes are and the position of the insertions in the alignment. It takes $\log_2 C_N^L = \log_2 \frac{L!}{(L-N)!N!}$ bits to encode the positions of $N$ insertions in an alignment of length $L$. In our example, the alignment length is 6 (the number of phonemes in the chosen lexical realization (5) plus the number of insertions (1)), yielding $\log_2 6$ bits. We complete the insertion portion of the message by adding to it information about the actual insertion performed. However, since we have already allocated space in the alignment for the insertion, it has turned into a substitution, which can be handled in the next step of the message sending process (it is now a "–" in the phoneme realization of the lexicon word, which is being substituted by an input phoneme).

**Sending the rest of the operations**. This is performed from left to right in the alignment. Since we know the top line of the optimal alignment, the bottom line is conveyed by sending a list of conditional probabilities of each input phoneme given the corresponding intended phoneme, which are computed from the training corpus, e.g., for "*another*$_2$" the probabilities are $P(\text{ax}|\text{ax})$, $P(\text{n}|\text{n})$, $P(-|\text{ah})$, $P(\text{dx}|\text{dh})$, $P(\text{q}|-)$ and $P(\text{er}|\text{er})$. These probabilities correspond to two exact matches, a deletion, a substitution, an insertion and an exact match.

Exact matches are common and thus take only a few bits to encode. In this manner, an input that closely matches one of the phoneme realizations will be encoded in fewer bits than one that is very different.

### 4.2.3 Summary of Sentence Evaluation

In summary, given a sentence $W* = w_1, w_2, \ldots, w_n$ where a word $w_i$ has part-of-speech $PoS_i$, and a partition of the input phoneme sequence into $n$ segments $phs_1, \ldots, phs_n$, we are trying to minimize the following:

$$code_length(W*) = \sum_{i=1}^{n}\left[code_length(w_i, PoS_i) + code_length(phs_i, w_i)\right],$$

where $code_length(w_i, PoS_i)$ is the number of bits required to send $w_i$ and $PoS_i$ (estimated from the language model), and $code_length(phs_i, w_i)$ is the number of bits required to send $phs_i$ given $w_i$ (estimated from the phoneme realization difference):[2]

$$code_length(w_i, PoS_i) = -\log_2\left[P(w_i|PoS_i, w_{i-1}) \times P(PoS_i|PoS_{i-1}, PoS_{i-2})\right]$$

$$code_length(phs_i, w_i) = \min_{\text{realizations } j \text{ of } w_i}\left[code_length(phReal_{w_i,j}) + code_length(\text{insertions}_{phReal_{w_i,j}}) + code_length(\text{substitutions}_{phReal_{w_i,j}})\right],$$

where $phReal_{w_i,j}$ is the $j$th realization of lexicon word $w_i$,

$$code_length(phReal_{w_i,j}) = -\log_2 \frac{Freq_{w_i,j}}{\sum_j Freq_{w_i,j}}$$

$$code_length(\text{insertions}_{phReal_{w_i,j}}) = \log_2 C_{\text{number of insertions}}^{\text{length of alignment}}$$

$$code_length(\text{substitutions}_{phReal_{w_i,j}}) = \sum_{\text{phonemes in alignment}} -\log_2 P(\text{input phoneme}|\text{intended phoneme}).$$

The argument $j$ that yields $code_length(phs_i, w_i)$ is the realization $j$ of lexicon word $w_i$ which yields the shortest encoding for the phoneme segment $phs_i$.

## 4.3 SEARCH THROUGH POSSIBLE SENTENCES

We have described a way to evaluate the quality of a sequence of word hypotheses that describe our input phonemes. But how do we search through the possible space of words and word boundaries?

Evaluating all possible sets of words with all possible sets of word boundaries would be computationally prohibitive. Thus, we follow an "optimistic selection" principle to eliminate partial sentences that are deemed unpromising during processing. This principle is implemented as follows. Each time a word is added to the current candidate partial sentences, those that take many more bits to encode than the best partial sentence at this stage of processing are eliminated from further consideration. This is because the partial sentences which take more bits to encode are unlikely to overcome this disadvantage and proceed to become the overall best when the whole sentence is eventually generated.

In our implementation we use a modified version of the Level-building algorithm, which proposes word boundaries and expands all partial sentence hypotheses a word at a time (Myers and Rabiner, 1981). At the end of each "level" (new word) we prune the partial sentence hypotheses using a beam threshold (Lowerre and Reddy, 1980). This allows us to have a strict control over the total number of partial hypotheses during the search. The phoneme slots that are generated by the algorithm are filled with suitably sized words, so that a slot with a few phonemes is not filled with a long word and a slot with many phonemes is not filled with a short word. The resulting words are then assigned different part-of-speech tags, and evaluated with each tag as described in Section 4.2.

To evaluate a particular word, optimal alignments are carried out between its phoneme realizations and the input phonemes. This process is quick due to the small number of phonemes in most words. However, the evaluation of all the words in the lexicon (comprising thousands of words) for a given phoneme string segment is a time consuming process. To reduce the number of word candidates to be evaluated at each point, we generate a short-list of likely possibilities. This is achieved by first encoding the phoneme realizations of words obtained from training as broad sound group sequences (e.g., the input sequence "dh ax r", which is a realization of "there", may be encoded as "stop vowel glide") and classifying each of these sequences into *Equivalence Classes*, such that each class contains phoneme realizations that are similar to each other, but each class as a whole is different from every other. The number and composition of these classes are optimized using the MML principle (Thomas *et al.*, 1996), yielding 38 classes with the current TIMIT data. During the search, a new string of input phonemes is placed in the "best" class according to a similarity measure which compares the sequences of broad sound groups corresponding to the input phonemes to those in each class. The words that correspond to the phoneme realizations in the chosen class become the short-listed candidates to be evaluated as described in Section 4.2. A disadvantage of using the short list is that the input phonemes may be put in a class that does not contain an instance of the correct word, and therefore the correct word will not be evaluated. Thus, the use of the short-list increases speed of recognition at the cost of losing some accuracy.

[2] The term $code_length(phs_i, w_i)$ is used instead of $code_length(phs_i|w_i)$ because it is a computational function call.

Table 1: Breakdown of the message length for two sentences from the TIMIT test set.

| Best Sentence | | | | | Correct Sentence | | | | |
|---|---|---|---|---|---|---|---|---|---|
| word | PoS bits | word bits | phDiff bits | total | word | PoS bits | word bits | phDiff bits | total |
| the | dt 4.22 | 0.94 | 3.18 | 8.34 | the | dt 4.22 | 0.94 | 3.18 | 8.34 |
| bank | nnp 5.45 | 8.74 | 18.86 | 33.05 | – | | | | |
| low | nnp 2.76 | 13.40 | 4.80 | 20.96 | bungalow | nn 1.84 | 14.83 | 17.88 | 34.55 |
| was | vbd 3.72 | 4.75 | 3.37 | 11.84 | was | vbn 4.10 | 3.16 | 3.37 | 10.63 |
| pleasantly | rb 2.76 | 11.12 | 25.17 | 39.05 | pleasantly | rb 2.88 | 11.12 | 25.17 | 39.17 |
| situated | vbn 2.98 | 2.32 | 35.83 | 41.13 | situated | vbn 2.98 | 2.32 | 35.83 | 41.13 |
| near | in 1.73 | 4.55 | 6.25 | 12.53 | near | in 1.73 | 4.55 | 6.25 | 12.53 |
| the | dt 1.41 | 0.26 | 2.87 | 4.54 | the | dt 1.41 | 0.26 | 2.87 | 4.54 |
| shore | nn 1.37 | 9.17 | 0.83 | 11.37 | shore | nn 1.37 | 9.17 | 0.83 | 11.37 |
| | 26.40 | 55.25 | 101.16 | 182.81 | | 20.53 | 46.35 | 95.38 | 162.26 |

Table 2: Phoneme realizations seen in training, for best and correct words of the first example of Table 1.

| Correct sentence phoneme realizations | *the*<br>dh ax<br>dh ix<br>dh iy | *bungalow*<br>b ah ng g ax l ow | | *was*<br>w ax z<br>w ix z<br>w ah z | *pleasantly*<br>p l eh z en t l iy | *situated*<br>s ih ch uw ey t ix d | *near*<br>n ih axr<br>n ih r<br>n ih er | *the*<br>dh ax<br>dh ix<br>dh iy | *shore*<br>sh ao r |
|---|---|---|---|---|---|---|---|---|---|
| Input Phonemes | dh ax | b ah ng g | el ow | w ah z | p l ah z en l ix | s ix ch ax-h w ix dx ih d | n ih | dh ix | sh ao r |
| Best sentence realizations | | *bank*<br>b ae ng k<br>b ay ng k | *low*<br>l ow | | | | | | |

## 5 RESULTS AND DISCUSSION

Table 1 shows the evaluation of a sample sentence from TIMIT's test set. We show the best sentence hypothesis after search, and also the sentence with the correct words and word boundaries for comparison. The table also shows the breakdown of the message into the number of bits required to encode the part-of-speech $N$-grams, the word $N$-grams given the part-of-speech symbol and the phoneme differences. A relatively large part-of-speech cost indicates an $N$-gram of low probability, but that cost may be offset by a relatively small cost to encode the word given that part-of-speech symbol or the phonemes given our chosen word. When the correct sentence requires fewer bits than the sentence found through search (as is the case with the example in Table 1), this indicates that the search was unable to find the optimal (correct) solution – the correct hypothesis was either not suggested or pruned out early. The converse suggests that the sentence actually found was a better hypothesis (a closer match between the phonemes, or a more likely word or phoneme sequence according to the language model) than the correct sentence. This could be due to the correct sentence being unusual and badly mismatched to data from training. An unusual sentence needs to have a good match between input phonemes and phoneme realizations of lexicon words in order to compensate for the low probability given by the language model. Table 2 shows the sample sentence of Table 1 in further detail, showing the input phoneme sequence, and some of the phoneme realizations of words for both the sentence found through search and the correct sentence. This example illustrates that it can be very difficult to distinguish between words that are phonetically similar given possible errors in the input; "bungalow" matches the input phonemes better than "bank low". However, the search did not suggest "bungalow" as a possible word candidate as it didn't belong to the shortlist generated for the input phonemes.

The left-hand side of Table 3 shows the results of the recognition of words in the standard core TIMIT test set, which contains 192 different sentences from 24 different speakers. None of the sentences were previously seen in training and none of the speakers were used in the training set.

Table 3: Average word error rates per sentence for differing levels of distortion, using word encoding.

| Distortion threshold | Average number of words | Part-of-Speech and Word Encoding | | | | Word Encoding | | | |
|---|---|---|---|---|---|---|---|---|---|
| | | average number of ins | average number of dels | average number of subs | average word error rate | average number of ins | average number of dels | average number of subs | average word error rate |
| 10 | 8.26 | 0.30 | 0.29 | 1.44 | 24.53 | 0.55 | 0.31 | 1.73 | 31.29 |
| 20 | 8.21 | 0.34 | 0.26 | 1.54 | 26.15 | 0.59 | 0.26 | 1.85 | 32.85 |
| 30 | 8.19 | 0.38 | 0.26 | 1.68 | 28.46 | 0.70 | 0.26 | 1.96 | 35.57 |
| 40 | 8.16 | 0.41 | 0.26 | 1.72 | 29.25 | 0.73 | 0.25 | 2.00 | 36.59 |
| 50 | 8.12 | 0.43 | 0.25 | 1.71 | 29.60 | 0.76 | 0.25 | 2.00 | 37.03 |
| 60 | 8.15 | 0.44 | 0.25 | 1.71 | 29.41 | 0.76 | 0.24 | 1.99 | 36.70 |

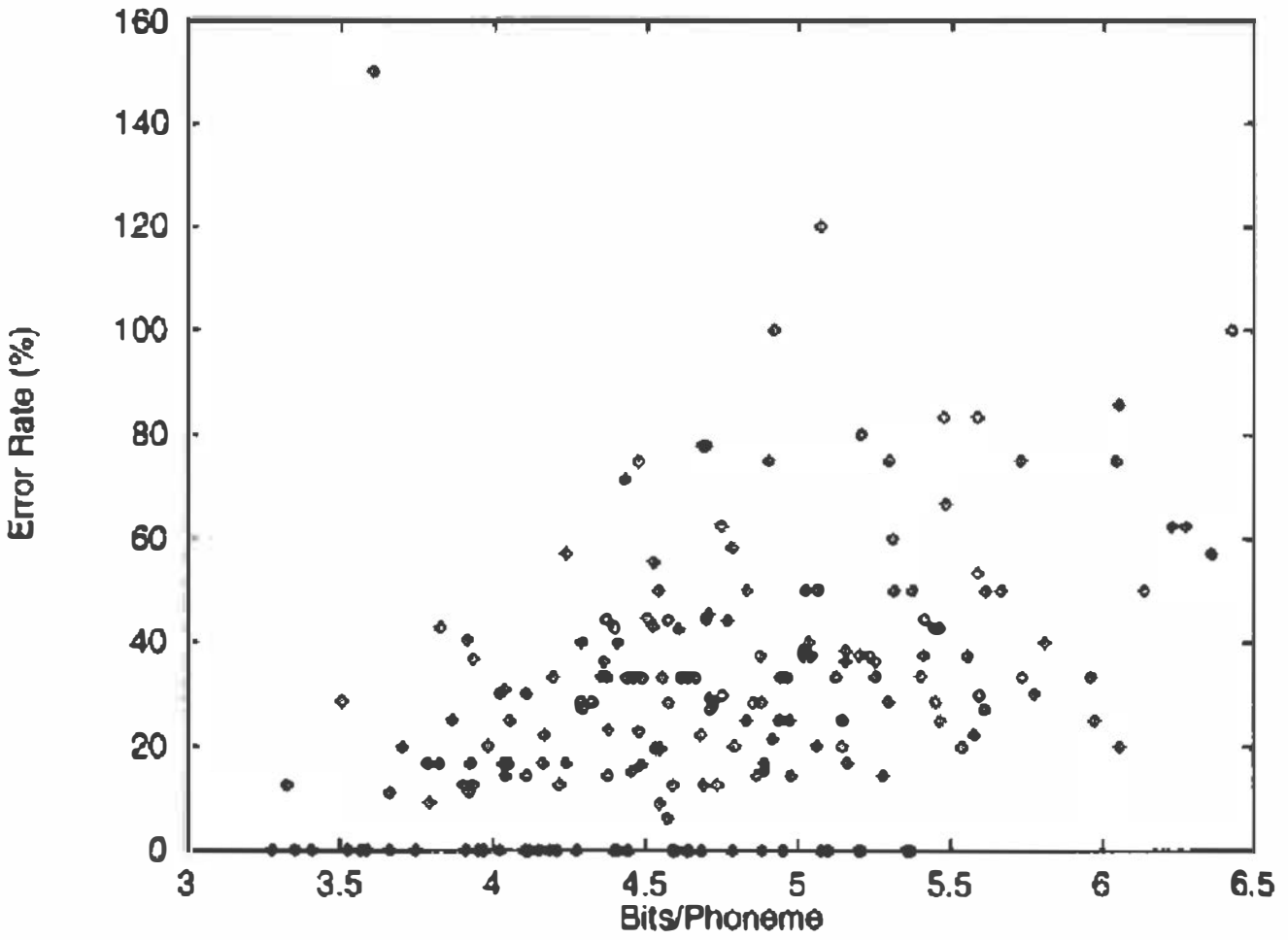

Figure 2: Error rate for bits/phoneme of 192 sentences.

The table shows the average number of words in the input sentences, and the average number of word insertions, deletions and non-exact substitutions in the alignment of the words in each of these sentences and the corresponding correct sentence. This is followed by the average word error rate, which is the sum of word insertions, deletions and non-exact substitutions divided by the true number of words. We further categorize these results according to the *Distortion Rate*. This is an estimate of how closely an input sequence of phonemes matches the sequence of phonemes for the corresponding correct sentence. It is calculated by first extracting the phoneme sequences that make up the correct words in the input sentence according to the correct word boundaries, and then finding the optimal alignment of each of the phoneme realizations of each correct word with the input phoneme sequence for that word. The number of insertions, deletions and substitutions in the closest alignment of a phoneme realization and an input phoneme sequence is recorded for each of the words in the sentence. The distortion rate for a sentence is the ratio of the number of non-exact phoneme matches (insertions, deletions and non-exact substitutions) to the total number of phonemes in the input phoneme sequence. For example, the distortion rate for the sample sentence in Table 2 is calculated as 11 non-exact operations (2 from "bungalow", 3 from "pleasantly", 5 from "situated" and 1 from "near") divided by 34 phonemes in the input sequence, giving a distortion rate of 32.35%. In Table 3, each line shows the average word error rate for all sentences which have a lower distortion rate than the value in the first column (e.g., for all sentences with less than 30% distortion, the word error rate was 28.46%).

In order to determine the effect that part-of-speech symbols have on the results, we analyzed the same set of sentences using a language model that contains word bigrams only (right-hand side of Table 3). The language model was calculated using the formula

$$P(W*) = \prod_{i=1}^{n} P(w_i|w_{i-1}).$$

Comparing the left and right sides of Table 3 we can see the improvement in the recognition accuracy when part-of-speech information is incorporated in the language model.

Figure 2 is a plot of the bits per phoneme versus the word error rate for each of the 192 sentences. We can see that as the number of bits per phoneme increases so too does the error rate. This would allow a recognizer to assess the likely accuracy of a found sentence – a high number of bits per phoneme would suggest a high error rate.

At present, the average word error rate is 29%. This recognition accuracy could be improved by careful weighting of the probabilities so that infrequent words still have a chance of being accepted if their phoneme realization is close to the input phonemes, or by using a more advanced method for modeling co-articulation between words than the one currently used (not described here due to space limitations). Other features of the signal such as phoneme duration can be also included as a further source of information.

Finally, the input to our system is a set of hand-marked phonemes, with no indication of the confidence of the markings. In a realistic system, the input would be a lattice of phoneme candidates and their probabilities. Such information could be factored into the encoding of a sentence; for example, if we have a high degree of confidence in a particular phoneme, then this might reduce the probability of non-matching substitutions involving this phoneme in the optimal alignment. Receiving as input a phoneme lattice would also allow us to compare our results with those obtained by other systems, e.g., (Rabiner and Juang, 1993).

## 6 CONCLUSION

We have shown the utility of Minimum Message Length Encoding for modelling spoken sentences. As with most corpus-based learning schemes, the quality of the training data is important. Given the diverse speakers and sentences involved, the recognition results are encouraging.

Part-of-speech data generally improved the number of bits required to send phoneme information. More importantly, it produced a shorter encoding of word hypotheses that were sensible in English, resulting in fewer word errors.

There was some concern over the lack of training data for the language model. This problem was reduced by using a larger body of classic texts, which supplied most common word $N$-grams, and more importantly, yielded a set of part-of-speech $N$-grams used commonly in English.

The use of short-lists of word candidates was shown to improve substantially the recognition speed of the system by quickly finding likely candidates to investigate more carefully, with minimal effect on recognition accuracy.

We are considering performance enhancements to the search, and also the evaluation of sentences, which is a major bottleneck of the system. Specifically, we are investigating a method for exploiting the optimistic selection principle during the evaluation of single words as well as partial sentence hypotheses. This allows a number of word evaluations to be carried out in lockstep, and allows the elimination of unlikely word candidates during evaluation.

**Acknowledgments**

This research was supported in part by grant N016/099 from the Australian Telecommunications and Electronics Research Board, and ARC Fellowship F39340111.

## References

Baum, L.E., An Inequality and Associated Maximization Technique in Statistical Estimation of Probabilistic Functions of Markov Processes. *Inequalities* 3, 1-8, 1972.

Béchet, F., Méloni, H., and Gillies, P., Knowledge-based Lexical Filtering: The Lexical Module of the SPEX System. In *Proc. of the Fifth Australian International Conference on Speech Science and Technology*, 528-532, Perth, Australia.

Bouchaffra, D., Koontz, E., Kṛpāsundar, V., Śrihari, R.K., and Śrihari, S.N., Integrating Signal and Language Context to Improve Handwritten Phrase Recognition: Alternative Approaches. In *Proc. of the Sixth International Workshop on AI & Statistics*, 47-54, Fort Lauderdale, Florida, 1997.

Fisher, W.M., Doddington, M., George, R., and Goudie-Marshall, K.M., The DARPA Speech Recognition Database: Specifications and Status. In *Proc. of the DARPA Speech Recognition Workshop*, Report No. SAIC-86/1546.

Fissore, L., Micca, G., and Pieraccini, R., Strategies for Lexical Access to Very Large Vocabularies. *Speech Communication* 7, 355-366, 1988.

Gauvain, J.L., Lamel, L., and Adda-Decker, M., Developments in Continuous Speech Dictation using the ARPA WSJ Task. In *ICASSP – Proc. of the IEEE International Conference on Acoustics, Speech and Signal Processing*, 65-68, 1995.

Grayden, D.B. and Scordilis, M.S., Phonemic Segmentation of Fluent Speech. In *ICASSP – Proc. of the IEEE International Conference on Acoustics, Speech and Signal Processing*, I-73-6, 1994.

Jeanrenaud, P., Eide, E., Chaudhari, U., McDonough, J., Ng, K., Siu, M., and Gish, H., Reducing Word Error on Conversational Speech from the Switchboard Corpus. In *ICASSP – Proc. of the IEEE International Conference on Acoustics, Speech and Signal Processing*, 53-56, 1995.

Lee, C. and Rabiner, L.R., A Frame-Synchronous Network Search Algorithm for Connected Word Recognition. *IEEE Transactions on Acoustics, Speech and Signal Processing* **37**(11), 1649-1658.

Lowerre, B. and Reddy, R., The Harpy Speech Understanding System. In *Trends in Speech Recognition*, Lea, W.A. (Ed.), Prentice-Hall, 340-360, 1980.

Murveit, H., Weintraub, M., Cohen, M., Bernstein, J., and Rudnicky, A., Lexical Access with Lattice Input. In *ICASSP – Proc. of the IEEE International Conference on Acoustics, Speech and Signal Processing*, 20.11.1-20.11.4, 1987.

Myers, C.S. and Rabiner, L.R., A Level Building Dynamic Time Warping Algorithm for Connected Word Recognition. *IEEE Transactions on Acoustics, Speech and Signal Processing* **29**(2), 284-297.

Oliver, J.J. and Baxter, R.A, MML and Bayesianism: Similarities and Differences. Technical Report 206, Department of Computer Science, Monash University. `http://www.cs.monash.edu.au/~jono`

Rabiner, L., and Juang, B, *Fundamentals of Speech Recognition*, Prentice Hall, 1993.

Sankoff, D. and Kruskal, J.B., *Time Warps, String Edits and Macromolecules: The Theory and Practice of Sequence Comparison*, Addison Wesley, London, 1983.

Thomas, I.E., Zukerman, I., and Raskutti, B., Accounting for Pronunciation of Phonemes in Corpora. In *Proc. of the Second Conference of the Pacific Association of Computational Linguistics*, forthcoming.

Thomas, I.E., Zukerman, I., Oliver, J.J., and Raskutti, B., Lexical Access using Minimum Message Length Encoding. In *PRICAI'96 – Proc. of the Fourth Pacific Rim International Conference on Artificial Intelligence*, Cairns, Australia, Springer-Verlag Berlin, 229-240, 1996.

Wallace, C.S. and Freeman, P.R., Estimation and Inference by Compact Coding. *Journal of the Royal Statistical Society (Series B)* **49**, 240-252, 1987.

Witten, I.H. and Bell, T.C., The Zero-Frequency Problem: Estimating the Probabilities of Novel Events in Adaptive Text Compression. *IEEE Transactions on Information Theory* **37**(4), 1085-1094.

Zue, V.W. and Lamel, L.M., An Expert Spectrogram Reader: A Knowledge-Based Approach to Speech Recognition. In *ICASSP – Proc. of the IEEE International Conference on Acoustics, Speech and Signal Processing*, 1197-1200, 1986.